\documentclass{article} 
\usepackage{iclr2023_conference,times}

\usepackage{bm, dsfont}

\newcommand{\bR}{\bm{R}}

\newcommand{\cB}{\mathcal{B}}

\newcommand{\cD}{\mathcal{D}}

\newcommand{\cL}{\mathcal{L}}

\newcommand{\cN}{\mathcal{N}}

\newcommand{\RR}{\mathbb{R}}

\makeatletter
\newcommand{\ve}{\@ifnextchar\bgroup{\velong}{{\bm{e}}}}
\newcommand{\velong}[1]{{\bm{#1}}}
\makeatother

\def\vtheta{{\bm{\theta}}}

\def\mE{{\bm{E}}}

\def\mH{{\bm{H}}}
\def\mI{{\bm{I}}}

\definecolor{citecolor}{HTML}{2779af}
\definecolor{linkcolor}{HTML}{c0392b}
\definecolor{urlcolor}{HTML}{c0392b}
\usepackage[pagebackref=false,breaklinks=true,colorlinks,bookmarks=false,citecolor=citecolor,linkcolor=linkcolor, urlcolor=urlcolor]{hyperref}

\usepackage{url}
\usepackage[ruled]{algorithm2e}
\usepackage{algpseudocode}
\usepackage{graphicx}
\usepackage{amsmath}
\usepackage{amsfonts} 
\usepackage{wrapfig}

\author{
Eric Zelikman \\
\texttt{\{ezelikman}
\And
Qian Huang \\
\texttt{qhwang}
\And
Percy Liang \\
\texttt{pliang}
\And
Nick Haber \\
\texttt{nhaber}
\And
Noah D. Goodman \\
\texttt{ngoodman\}@stanford.edu}
}

\iclrfinalcopy 
\begin{document}

\title{Just One Byte (per gradient): A Note on Low-Bandwidth Decentralized Language Model Finetuning Using Shared Randomness}

\maketitle
\vspace{-20px}

\begin{abstract}
Language model training in distributed settings is limited by the communication cost of gradient exchanges. In this short note, we extend recent work from \citet{malladi2023fine}, using shared randomness to perform distributed fine-tuning with low bandwidth. The method is a natural decentralized extension of memory-efficient Simultaneous Perturbation Stochastic Approximation (SPSA). Each iteration, each machine seeds a Random Number Generator (RNG) to perform local reproducible perturbations on model weights and calculate and exchange scalar projected gradients, which are then used to update each model. By using a (machine, sample) identifier as the random seed, each model can regenerate one another's perturbations. As machines only exchange single-byte projected gradients, this is highly communication efficient. There are also potential privacy benefits, as projected gradients may be calculated on different training data, and models never access the others' data.
Our approach not only drastically reduces communication bandwidth requirements but also accommodates dynamic addition or removal of machines during the training process and retains the memory-efficient and inference-only advantages of recent work. We perform proof-of-concept experiments to demonstrate the potential usefulness of this method, building off of rich literature on distributed optimization and memory-efficient training.\footnote{We release our code at \url{https://github.com/ezelikman/justonebyte}}
\end{abstract}

\vspace{-10px}
\section{Introduction}
\vspace{-5px}
More compute somewhat predictably results in better language models \citep{kaplan2020scaling}. This has motivated companies to spend hundreds of millions of dollars on large-scale language model training runs \citep{techcrunch2023anthropics}. On the other hand, distributed computing across many internet devices' spare compute has been used to discover new drugs, understand diseases, find pulsars, and model climate change  \citep{jayachandran2006parallelized,nasica2015amyloid,clark2015psr,stainforth2004climateprediction}. A natural next step is decentralized large language model training \citep{together2023gptjt}. However, this presents a challenge: as increasingly larger models run on consumer hardware, training them requires exchanging larger gradients. Some have suggested decentralized training on iPhones by training a subset of parameters \citep{dettmers2023qlora}. Yet, if doing full fine-tuning, uploading a 33-billion parameter gradient on a 5G Google Fi Flexible mobile plan \citep{GoogleFi2023} would take 14 hours and cost \$1,300 -- which some smartphone users may take issue with.\looseness=-1

Many works have highlighted that shared sources of randomness can improve communication efficiency \citep{theis2022lossy,canonne2015communication,acharya2019communication,kurri2021coordination,isik2022sparse}, as well as the need for randomness for privacy in federated learning \citep{beguier2020efficient,chen2022fundamental,ben2022scionfl}. In addition, numerous studies have centered the immense cost of communicating gradients in distributed training \citep{agarwal2022utility,recht2011hogwild}. Recent work from \citet{malladi2023fine} proposed a clever technique: by perturbing a language model's weights randomly and reverting those changes by reusing a random seed, one can calculate the loss of a perturbation using only forward passes and with practically no additional memory. They apply simultaneous perturbation stochastic approximation (SPSA) \citep{spall1992multivariate}: once the loss is calculated, the perturbation can be subtracted from the model parameters, weighted by the calculated scalar-valued projected-gradient. Thus, to perform each model update, all one needs is the projected gradient and the seed used to generate its corresponding perturbation.

So, we highlight an extension to \citet{malladi2023fine}'s memory-efficient zerothorder optimizer (MeZO): as each update ultimately needs only the scalar projected gradient and corresponding random seed, a network of machines can perform decentralized training by sharing only projected gradients. Since applying a random perturbation is computationally cheaper than a forward pass, forward passes can be easily parallelized. This allows distributed fine-tuning, where each machine holds an identical local model copy and needs only to communicate its projected gradients.
This has advantages beyond bandwidth: because only scalar projected gradients are shared, machines may train on different data. We also discuss potential privacy advantages in Appendix~\ref{privacy}. As each machine maintains a model copy, adding or removing machines is straightforward, unlike techniques designating subsets of data to each machine \citep{li2022branch}. Finally, like MeZO, this is low-memory and gradient-free, allowing training on non-differentiable objectives.

\begin{figure}[t]
    \centering
    \includegraphics[width=0.9\textwidth]{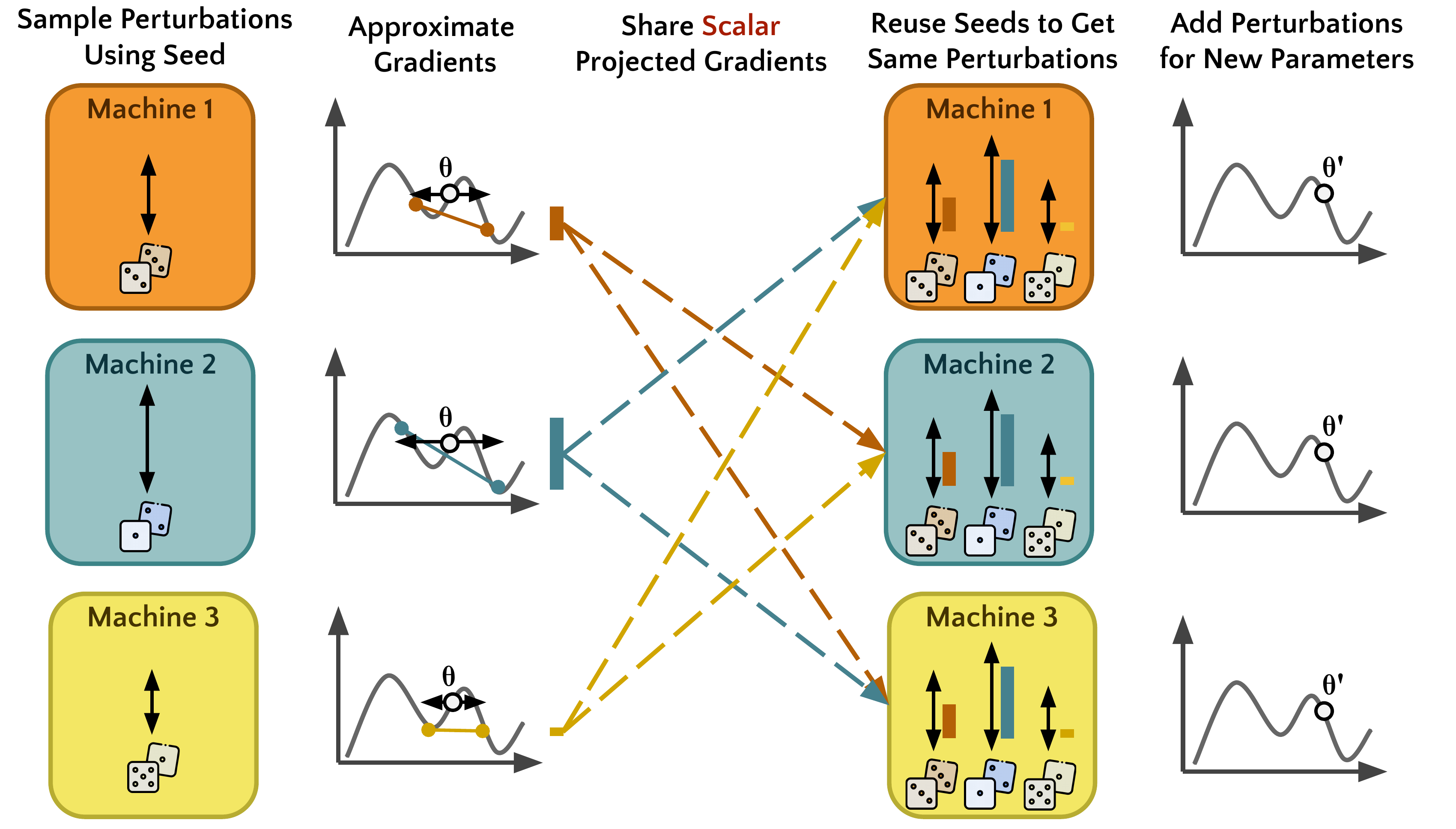}
    \vspace{-5px}
    \caption{Overview of our method. Each machine calculates the impact of a seeded random weight perturbation on the loss. The scalar-valued projected gradient is then shared with all other machines, which is used to update each machine's model by regenerating the perturbations.}
    \vspace{-10px}
    \label{fig:overview}
\end{figure}

\vspace{-5px}
\section{Method}
\vspace{-5px}

\textbf{Initialization}. Each machine connects to the network to receive the current training iteration and a list of known machines, and the existing machines notify their peers of the new machine.

\textbf{Evaluating deterministically-random perturbations.} The machine then begins an inference loop. Each inference includes multiple perturbations of the model parameters based on a random seed $s$, which is constructed from the machine identifier time, the current iteration, and the index of the current sample. As in \citet{malladi2023fine}, in each inference, we calculate the scalar \textit{projected gradient}, the change in loss as a perturbation is added and subtracted. The iteration ends when all active machines have completed the desired number of inferences or when the timeout is reached.

\textbf{Applying collected scalar projected gradients.} The projected gradients are aggregated across all machines. Once all of the gradients have been received, each machine constructs a dictionary mapping a machine timestamp to a list of that machine's calculated scalar projected gradients; the machines then apply the full set of gradients. Specifically, in machine-timestamp order, each machine applies the other perturbations using the same random seed $s$ used to generate the perturbation, weighted by the projected gradient and negative learning rate. We also (optionally) back up the weights to memory every $n$ inference loops and restore it before applying the collected gradients to avoid the impact of any floating point errors from perturbations (i.e., $\vtheta$ + $z$ - $z$ != $\vtheta$).

\textbf{Synchronizing after gradient application.} Each machine informs every other machine that it is ready to proceed and waits to confirm that every machine has applied its gradients -- any machine that does not respond must reconnect. Each machine checks with a random peer whether they have the same model checksum. Then, the global number of iterations increases and the training continues.

\vspace{-5px}
\section{Experiments} \vspace{-5px}
Our primary objective was to validate that the proposed method works as expected, focusing on its communication bandwidth and impact on decentralized training speed. The accuracy of the perturbation-based SPSA for fine-tuning language models were recently confirmed in a study by \citet{malladi2023fine}, and our method builds on this work. As they highlight in their Appendix on ``Sample Schedules,'' sampling more projected gradients per timestamp does not harm performance, even with the same total number of forward passes. In other words, they found that fine-tuning with this approach is not harmed by parallelization -- at least, up to the 16 parallel samples they considered. As our primary focus is on the distributed performance of this algorithm, and the effectiveness of SPSA for fine-tuning language models was demonstrated by \citet{malladi2023fine}, we select a dataset also used by them, SST2 \citep{socher2013recursive}, and focus our evaluations on it. We analyze a 6.7-billion parameter OPT model \citep{zhang2022opt}, trained for a total of 16,000 gradient steps with forward-only training and 2.5x fewer steps for standard training, following the hyperparameters from \citet{malladi2023fine}. We do not compare directly to low-rank optimization in our training evaluation but note that it inherently, significantly constrains the optimization of the model \citep{hu2021lora}.

\begin{wrapfigure}{r}{0.4\textwidth}
\vspace{-10px}
{
\centering
\includegraphics[width=0.4\textwidth]{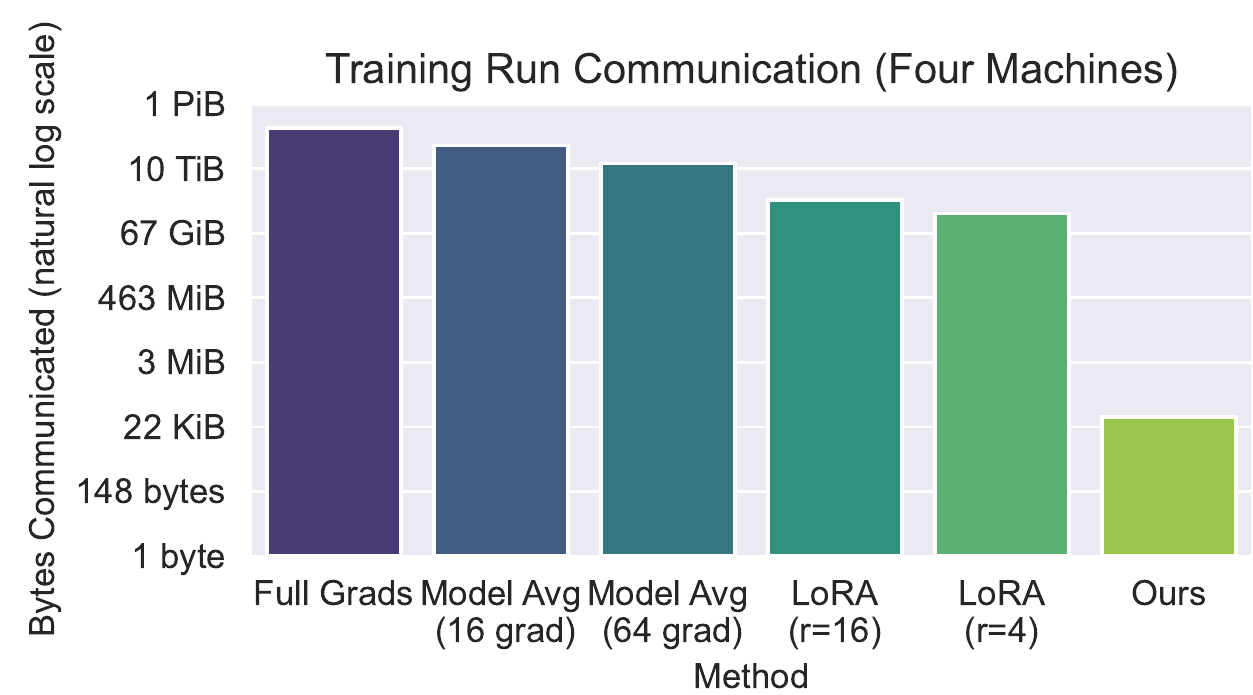}
}\vspace{-20px}
\caption{Overview of our proposed method's communication between four machines relative to other fine-tuning approaches with gradient accumulation.}
\label{fig:comm}
\vspace{-10px}
\end{wrapfigure}

\textbf{Communication Bandwidth:} We measured the amount of data exchanged by the machines during the training process in terms of the scalar projected gradients. We compare to a (very) naive baseline of sending all of the gradients between each machine used for training -- in fact, this baseline is so naive we are unable to operationalize it for the 6.7 billion parameter OPT model on which we perform most of our analyses, even when using 16-bit bfloat precision. As a result, we perform our full-gradient analysis on a single machine and calculate the communication that would have been necessary: roughly 250 terabytes to parallelize the training over four machines. We note a slightly less naive but computationally equivalent approach of model averaging after every iteration would have required a mere 16 terabytes for 64-gradient accumulation (16 per machine) \citep{guo2021model}. Less naive communication methods like communicating the gradients of a low-rank adapter \citep{hu2021lora} are also visualized in Figure~\ref{fig:comm}.

\begin{figure}[t]
    \centering
    \vspace{-10px}
    \includegraphics[width=1.05\textwidth]{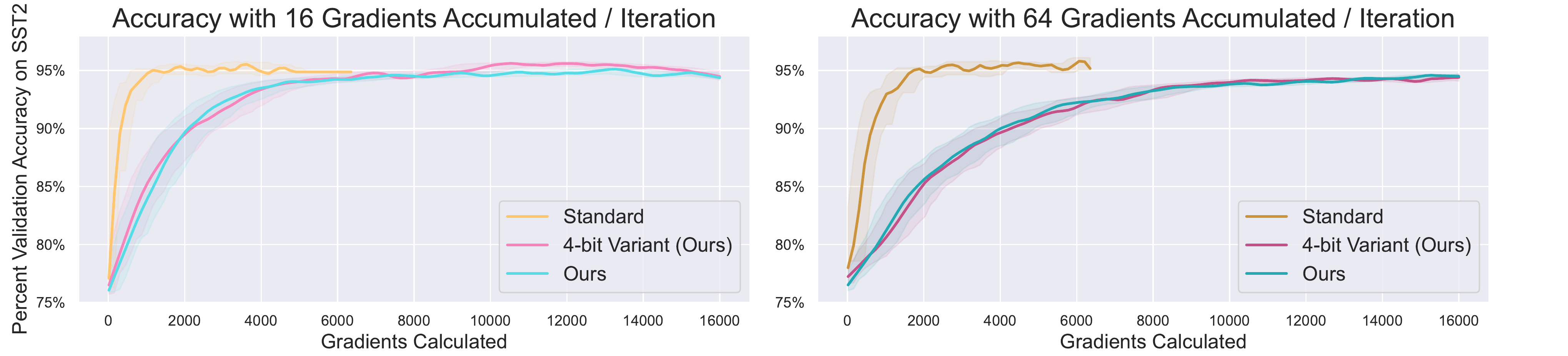}
    \vspace{-10px}
    \caption{Overview of our proposed method's performance distributed over four machines compared to standard fine-tuning with gradient accumulation with the same number of total computed gradients (i.e., total gradients computed combined across all machines).}
    \vspace{-10px}
    \label{fig:sstspeed}
\end{figure}

\textbf{Training Speed:} We monitored the number of projected gradients calculated by our approach as well as the number of gradients computed by traditional fine-tuning and plot them against one another. We visualize these results in Figure~\ref{fig:sstspeed}. We also explore 4-bit forward-pass-only training by dequantizing layer weights before adding noise and then re-quantizing them, finding no impact on performance \citep{dettmers2023qlora}. We emphasize that our goal is not to achieve the efficiency of gradient descent in terms of the number of gradients computed but to demonstrate that we can accomplish similar performance with minimal communication. In order to compress the projected gradients to a single byte, staying true to the title of this paper, we also test communicating the signed log of the gradient clipped to (-127, 127). We observe this has no measurable impact on performance compared to simply sending the floating point value of the projected gradient.

\vspace{-5px}
\section{Discussion}\vspace{-8px}
While we inherit the many benefits of MeZO, we also inherit their limitations \citep{malladi2023fine}. First, in terms of inferences needed, SPSA is generally far slower than gradient descent -- however, we have not explored tasks where the target is not differentiable and where optimization is traditionally done using reparameterization, such as in the REINFORCE algorithm or in the context of variational autoencoders (VAEs) \citep{zhang2021sample,kingma2013auto}. Since SPSA is indifferent to the differentiability of the target function, it is unclear whether SPSA would be competitive in these settings. Second, it is unclear which rules-of-thumb generalize to SPSA. For example, we scaled the learning rate with $\mathrm{sqrt}$ of the number of sampled perturbations, corresponding to the standard deviation of the sum of independent normal samples. Correspondingly, theoretical results suggest similar scaling properties with batch size in traditional neural network training \citep{krizhevsky2014one}. We discuss additional limitations in Appendix~\ref{limitations}.

\vspace{-5px}
\section{Related Works}\vspace{-10px}
We note that many prior works leverage randomness for communication efficiency \citep{theis2022lossy,canonne2015communication,acharya2019communication,kurri2021coordination,isik2022sparse} and privacy in federated learning \citep{beguier2020efficient,chen2022fundamental,ben2022scionfl,ayle2023training}, although distributed SPSA \citep{ramaswamy2017analysis,ramaswamy2020dspg,sergeenko2021convergence} has not incorporated this. 
Challenges in decentralized training include \textit{varying processing abilities} \citep{luo2019hop}, which we address by supporting count-based and timeout-based systems for training iterations. Research also covers \textit{data heterogeneity}, with methods proposed to consider intra- and inter-machine data variance \citep{tang2018d} -- though our approach supports multiple data sources, further analysis is needed to evaluate its impact. Other training works also address communication; e.g., \citet{tang2018communication} and \citet{koloskova2019decentralized} improve bandwidth by aggregating communications from peers while \citet{yuan2022decentralized} decompose training into geographical "tasklets" to minimize communication, and \citet{recht2011hogwild} trains without synchronization. Lastly, there is also a body of work on reducing the impact of adversarial machines in decentralized training that do not behave as expected \citep{elkordy2022basil,kuwaranancharoen2023geometric}. While an important direction, we assume a trusted network within this work.

SPSA \citep{spall1992multivariate} is an extensively studied zeroth-order optimization algorithm \citep{gerencser1999optimization,tympakianaki2015c,maryak1999efficient,tympakianaki2018robust,glick2021covariant,gacon2021simultaneous,wiedmann2023empirical} and underlies \citet{malladi2023fine}.
There are also layerwise SPSA variants \citep{wulff2018spsa,tacchino2021variational} which may be possible to extend allow for further decentralization of the model, allowing subsets of layers to be stored on different machines while increasing the communication cost only on the order of the number of layers in the network, as well as techniques to efficiently estimate the Hessian and perform second-order optimization \citep{zhu2002modified}. Recent work has highlighted the potential impact of second-order optimization for fast language model pretraining, so incorporating these optimizations into our algorithm would be a natural next step \citep{liu2023sophia}.

\vspace{-5px}
\section{Conclusion}
\vspace{-10px}
In this paper, we presented a low-bandwidth, decentralized language model fine-tuning approach that leverages shared randomness, inspired by the work of \citet{malladi2023fine}. Our method involves machines performing parallel inferences with randomly perturbed weights and sharing only the scalar projected gradients. This technique reduces communication costs, offers potential privacy enhancement, and allows adding or removing machines. We evaluated our approach in terms of communication bandwidth and training speed, showing significant reduction in bandwidth while maintaining competitive model performance.

However, several areas warrant further investigation, including addressing the limitations of SPSA and analyzing the impacts of training on multiple data sources, understanding the behavior under adversarial circumstances, and further optimizing strategies for large-scale mesh networks. In addition, an exploration of the efficacy of this approach in pretraining would be valuable. However, this may require incorporating more ideas from SPSA literature, such as global annealing and automatic learning rate tuning \citep{yuan2008model}. In summary, our work provides a step on a path towards more efficient, scalable, and private decentralized training of large language models.

\subsection*{Acknowledgments}
We thank Simran Arora, Fan-Yun Sun, Violet Xiang, and Logan Cross for their helpful feedback on this work.

\bibliography{main}
\bibliographystyle{iclr2023_conference}

\vfill
\pagebreak
\appendix
\section{Algorithm}
\begin{algorithm}[h]
\small
\SetAlgoNoEnd
\SetKwFunction{perturb}{PerturbParameters}
\SetKwProg{sub}{Subroutine}{}{}
\SetKwFunction{sync}{SyncMachines}
\SetKwFunction{getGradients}{GetGradients}
\SetKwFunction{applyGradient}{ApplyGradient}
\SetKwFunction{checksum}{ChecksumMatch}
\SetKwFunction{connect}{ConnectToNetwork}
\SetKwFunction{sleep}{Sleep}

\textbf{Require}: initial parameters $\vtheta\in\RR^d$, 
loss $\cL:\RR^d\to\RR$, 
perturbation scale $\epsilon$, 
learning rate schedule $\{\eta_t\}$,
per-iter inferences $n_{\mathrm{inferences}}$,
timeout $T_{\mathrm{timeout}}$,
known machines $\mathbf{M}$,
max iterations $max\_iter$,
start time $self\_time$, gradient application timeout $T_{\mathrm{apply\_grads}}$\\
\vspace{0.2cm}
$cur\_iter, \mathbf{M} \gets \connect(self\_time$, $\mathbf{M}$)\\
\While{$cur\_iter < max\_iter$}{
$T_{\mathrm{start}} = $ time.time() \\
$\vtheta_{\mathrm{backup}} \gets \mathrm{copy}(\vtheta)$\\
$synced = \mathrm{defaultdict}(\mathrm{set})$\\
\While{\textnormal{time.time()} $\leq T_{\mathrm{start}} + T_{\mathrm{timeout}}$ and $\mathbf{M} \neq synced[\mathrm{`performed\_inferences'}]$}{
    Sample batch $\cB\subset \cD$\\
    $s \gets ({self\_time}, t, \mathrm{length(\texttt{projected\_grads}}))$ \\
    $\vtheta\gets$ \perturb{$\vtheta, \epsilon, s$} \\
    $\ell_+\gets\cL(\vtheta;\cB)$ \\
    $\vtheta\gets$ \perturb{$\vtheta, -2\epsilon, s$} \\
    $\ell_-\gets\cL(\vtheta;\cB)$ \\
    $\vtheta\gets$ \perturb{$\vtheta, \epsilon, s$} \\
    $\textnormal{\texttt{projected\_grads}}$.append($(\ell_+ - \ell_-) / (2\epsilon)$) \\
    \If{$\mathrm{length}$($\textnormal{\texttt{projected\_grads}}$) == $n_{\mathrm{inferences}}$}{
        \sync{`performed\_inferences'}
    }
}
machine\_grads = \getGradients{$\mathbf{M}$} \Comment{Timestamp-ordered dictionary} \\
$\vtheta \gets \mathrm{copy}(\vtheta_{\mathrm{backup}})$ \Comment{Mitigate drift due to floating point errors}\\
\For{timestamp, grads $\in$ all\_grads.items()}{
    \For{sample\_id, grad $\in$ enumerate(grads)}{
        $s \gets$ (timestamp, cur\_iter, sample\_id) \\
        \perturb{$\vtheta, -grad * \eta / numel(all\_grads), s$} \\
    }
}
$cur\_iter  \gets  cur\_iter$ + 1\\
\sync{`applied\_gradients', blocking=True, timeout=$T_{\mathrm{apply\_grads}}$} \\
\checksum{random.choice($\mathbf{M}$)} \Comment{Check if hashed $\vtheta$ matches peer's} \\
}
\vspace{0.2cm}
\sub{\perturb{$\vtheta$, $\epsilon$, $s$}}{
    Reset random number generator with seed $s$ \\
    \For{$\theta_i\in\vtheta$ }{  
        $z\sim\cN(0,1)$ \\
        $\theta_i\gets\theta_i+\epsilon z$
    }
    \Return $\vtheta$
}
\vspace{0.2cm}
\sub{\sync{$sync\_param$, $blocking$, $timeout$}}{
$m.synced[sync\_param]$.add($self\_time$) \textbf{for} $m\in M$ 
\Comment{Via network call}\\
\If{$blocking$}{
$start$ = time.time() \\
\textbf{while} {$start + timeout > $ time.time() and $\mathbf{M} \neq synced[sync\_param]$}{ \textbf{do}
\sleep{} \\
}
$\mathbf{M}$ = $synced[sync\_param]$ \Comment{Remove disconnected machines}
}
}
\caption{Distributed Training with Just One Byte (Modified from \citet{malladi2023fine}}
\label{alg:dist_train}
\end{algorithm}

\vfill
\newpage
\section{Further Limitations}
\label{limitations}
In the main text, we highlighted some limitations of this method -- here, we note some additional limitations. Another limitation is that, for models to be able to join after the first iteration, they need to be able to receive the model weights within a single iteration. This is naturally less expensive than sending the full set of gradient updates every step, but for very large models it may still be a very large amount of data for a single low-bandwidth machine to communicate. One potentially useful optimization here would be for the new machine to gather the weights from multiple peers, reducing the burden on any single machine and allowing a high-bandwidth machine to download weights more quickly. Yet, there is another limitation here: if the time between each iteration is less than the time it takes to download the weights, then the new machine will never be able to catch up. There is a natural resolution to this: once the weights are downloaded, the new machine can apply all of the new projected gradients from any iterations that have passed while it was downloading the weights. This would allow the new machine to catch up to the rest of the cluster, but it would also require either 1) the machine to query the other machines for the projected gradients while it downloads the weights or 2) the machines providing weights to keep track of the past projected gradients they have computed. Both come with tradeoffs: the first is potentially fragile, as any interruption in the download would require the machine to restart the download and re-query the other machines, while the second requires the machines to store additional data.

One key challenge is the underlying assumption the ability to reproduce noise deterministically across different machines, does not necessarily hold across different GPU architectures, not to mention other kinds of computing devices like TPUs and CPUs. For example, PyTorch uses a built-in CUDA random number generator (for Nvidia devices). This is deterministic with multiple uses of the same random seed on one machine, but different Nvidia GPU architectures implement this random noise differently. As a result, GPU-generated noise in PyTorch cannot be deterministic across machine types \citep{antoche2022}. On the other hand, JAX places an emphasis on reproducibility, and its random number generator is deterministic across different machine types. However, JAX's random number generator is multiple orders of magnitude slower than PyTorch's. Because we apply noise to the weights of the model, this means that the cost of noise generation is proportional to the number of parameters in the model. If one knows that their entire network will be composed of machines with the same machine learning accelerators, there is no obvious best choice. While our primary implementation is in PyTorch, we have also implemented a version in JAX for these benefits. A related challenge is the limitations of floating point precision - even applying identical floating point transformations on two different machines can result in different floating point outputs. IEEE 754, which is the standard for floating-point arithmetic, supports multiple rounding modes, has multiple revisions, and different handling of numbers that are close to zero (denormal numbers) \citep{ieee2019ieee}.

Finally, while our method inherently communicates less information than a full gradient update, it is difficult to quantify the exact privacy gains. For a particular weight and a particular batch of data, as $\epsilon$ approaches 0 and the number of sampled gradients approaches infinity, we would expect to perfectly replicate the gradient. The relative amount of usable information conveyed by each projected gradient must be greater than its relative size in bytes (or a model would take centuries to fine-tune to comparable performance). A deeper, more formal analysis of the privacy implications of our method would be absolutely essential before wide-scale deployment in contexts where privacy is needed. One hope would be that, with sufficient privacy, this method could theoretically be employed for continual learning by leveraging spare compute on edge devices. Indeed, a slightly modified version of this algorithm, where the model uses only a single perturbed inference and tracks its previous loss, would require no additional compute to train than the inference itself.

\section{Preliminary Privacy Implications}
\label{privacy}
It is not straightforward to estimate the privacy implications of this method. On one hand, extensive work suggests that various gradient compression methods, of which this algorithm can potentially be seen as an extreme version, can substantially improve privacy \citep{melas2022intrinsic,wang2021datalens,huang2020privacy}. On the other hand, prior work has shown that the use of shared sources of randomness can allow for more information to be communicated \citep{theis2021advantages,theis2022lossy}. However, if we assume a prior on the gradient of a multivariate normal (that is $\nabla \cL(\vtheta) \sim \cN(c, \Sigma)$ where $c\in\bR^k, \Sigma\in\bR^{k\times k}$), then because a projected gradient acts as a one-dimensional reduction in the possible space of gradients, we can derive the expected change in entropy: 
\begin{align}
\mE_v[\mH(\nabla \cL(\vtheta)) - \mH(P_v \nabla \cL(\vtheta))]
\\ = \left(\frac{1}{2}\log(|\Sigma|) - \frac{k}{2}(1 + \log{2\pi})\right) - \left(\frac{1}{2}\log(|\Sigma'|) - \frac{k-1}{2}(1 + \log{2\pi})\right)
\\ = \frac{1}{2}\log(\frac{|\Sigma|}{|\Sigma'|}) + \frac{1 + \log{2\pi}}{2}
\end{align}
If we were to assume $\Sigma = \mI$, the proportion of the total entropy would be $\frac{\mH(\nabla \cL(\vtheta))}{k}$. More realistically, we have $\frac{\mH(\nabla \cL(\vtheta))}{k} \frac{1 + \log(2 \pi) + \log|\Sigma'| - \log|\Sigma|}{1 + \log(2 \pi) + \log|\Sigma| - \log k}$. Without an extremely high degree of covariance, we expect each projected gradient to represent only a very small fraction of the information of the true gradient in high dimensions.

\end{document}